\documentclass[conference]{IEEEtran}
\IEEEoverridecommandlockouts
\usepackage{url}   
\usepackage{cite}
\usepackage{amsmath,amssymb,amsfonts}
\usepackage{algorithmic}
\usepackage{graphicx}
\usepackage{textcomp}
\usepackage{xcolor}
\usepackage{subfig}
\usepackage{threeparttablex}
\usepackage{commath}

\def\BibTeX{{\rm B\kern-.05em{\sc i\kern-.025em b}\kern-.08em
    T\kern-.1667em\lower.7ex\hbox{E}\kern-.125emX}}
\begin{document}

\title{Recurrent Neural Networks based Obesity Status Prediction Using Activity Data \\
\thanks{}
}
\author{\IEEEauthorblockN{Qinghan Xue\IEEEauthorrefmark{1}\IEEEauthorrefmark{2},
Xiaoran Wang\IEEEauthorrefmark{1},
Samuel Meehan\IEEEauthorrefmark{1},
Jilong Kuang\IEEEauthorrefmark{1},
Alex Gao\IEEEauthorrefmark{1} and 
Mooi Choo Chuah\IEEEauthorrefmark{2}}
\IEEEauthorblockA{\IEEEauthorrefmark{1}Samsung Research America.
Mountain View, CA 94043\\ Email: x.wang@samsung.com; s.meehan@samsung.com; jilong.kuang@samsung.com; alex.gao@samsung.com}
\IEEEauthorblockA{\IEEEauthorrefmark{2}Department of Computer Science and Engineering, 
Lehigh University,
Bethlehem, PA 18015\\ Email: qix213@lehigh.edu; chuah@cse.lehigh.edu}}
%\author{\IEEEauthorblockN{1\textsuperscript{st} Qinghan Xue}
%\IEEEauthorblockA{\textit{Department of Computer Science and Engineering} \\
%\textit{Lehigh University}\\
%Bethlehem, PA, 18015 \\
%qix213@lehigh.edu}
%\and
%\IEEEauthorblockN{2\textsuperscript{nd} Xiaoran Wang}
%\and
%\IEEEauthorblockN{3\textsuperscript{rd} Samuel Meehan}
%\and
%\IEEEauthorblockN{4\textsuperscript{th} Jilong Kuang}

\maketitle

\begin{abstract}

Obesity is a serious public health concern worldwide, which increases the risk of many diseases, including hypertension, stroke, and type 2 diabetes. 
To tackle this problem, researchers across the health ecosystem are collecting diverse types of data, which includes biomedical, behavioral and activity, and utilizing machine learning techniques to mine hidden patterns for obesity status improvement prediction. While existing machine learning methods such as Recurrent Neural Networks (RNNs) can provide exceptional results, it is challenging to discover hidden patterns of the sequential data due to the irregular observation time instances. Meanwhile, the lack of understanding of why those learning models are effective also limits further improvements on their architectures. Thus, in this work, we develop a RNN based time-aware architecture to tackle the challenging problem of handling irregular observation times and relevant feature extractions from longitudinal patient records for obesity status improvement prediction. To improve the prediction performance, we train our model using two data sources: (i) electronic medical records containing information regarding lab tests, diagnoses, and demographics; (ii) continuous activity data collected from popular wearables. 
%In addition, we propose an analytical method to explain the functions of the hidden state units in our model based on their expected responses to input data. 
Evaluations of real-world data demonstrate that our proposed method can capture the underlying structures in users' time sequences with  irregularities, and achieve an accuracy of 77-86\% in predicting the obesity status improvement.

\end{abstract}
\begin{IEEEkeywords}
obesity surveillance, activity data, hidden patterns, sequential data, recurrent neural network
\end{IEEEkeywords}

%\input{intro.tex}
%\vspace{-0.06in}
\section{Introduction}
%\vspace{-0.06in}

In recent years, the advent and rapid adoption of mobile health (mHealth) \cite{bandodkar2016wearable,quisel2017collecting} enabled by wearable technologies have made continuous monitoring of environment and lifestyle a concrete possibility. 
For example, tiny low cost sensors such as smart watches and wristbands track a variety of parameters that range from steps taken and hours slept, to heart rate variability, which enable intelligent healthcare applications to provide estimates of steps, calories, fitness assessment, rehabilitation, activity duration, and may even offer disease pre-diagnosis \cite{alemdar2010wireless,caytiles2014study,filipe2015wireless}.

\vspace{-0.05in}
\begin{figure}[!htbp]
\centering \includegraphics[height=1.4in,width=3.4in]{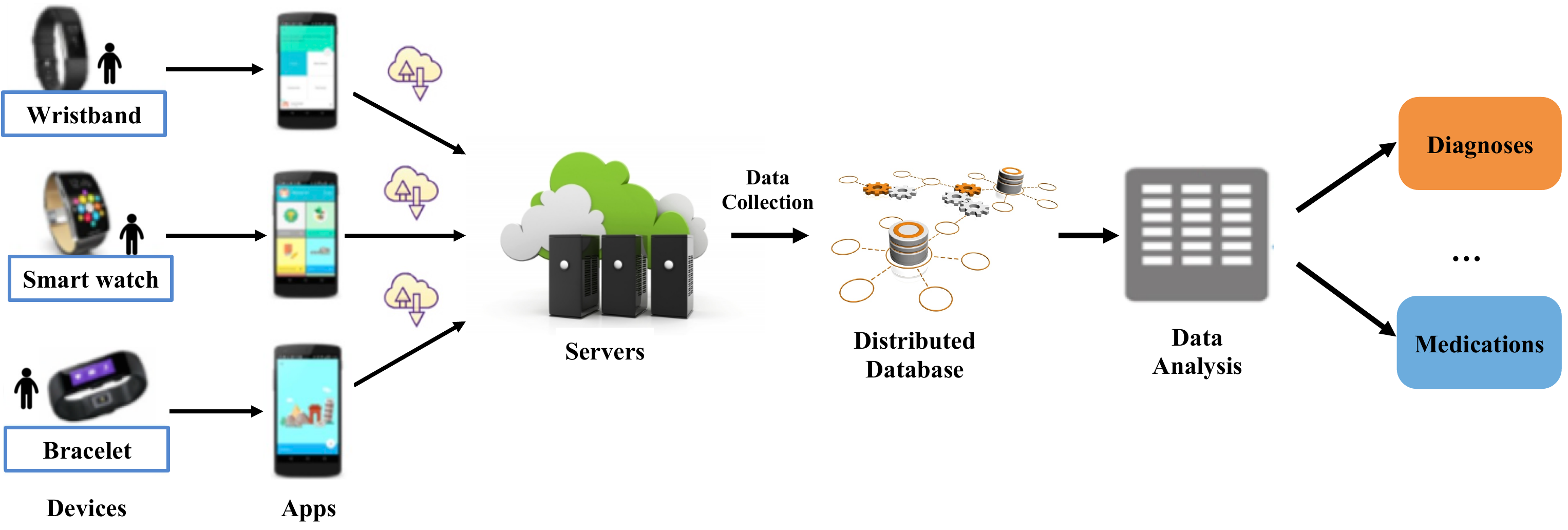}
\vspace{-0.15in}
\caption{\small End-to-end Data Flow}
\label{fig:intro-mod}
\vspace{-0.2in}
\end{figure}
%Thus, with the extensive use of sensors (e.g., wearable devices) large volumes of heterogeneous data streams have become available. 
Fig~\ref{fig:intro-mod} shows an example of a system where data streams are collected using various sensors/mobile devices and analyzed to understand users' physical activity levels, and possible disease forecasting. A particular disease of interest is obesity since it is a critical worldwide problem. Early detection and status monitoring of such patients allow better care.
%For example, it is essential for early detection of the obesity disease which has become a critical problem in United States and worldwide so that appropriate interventions can be taken to prevent people from developing critical diseases associated with obesity. 

While studies have shown that machine learning methods can provide good performance in various healthcare applications for personalized disease diagnosis, medication, and treatments, it is still challenging to learn efficient patterns from heterogeneous healthcare data.
To overcome this, recently, deep learning techniques (e.g., RNN) have been adopted in medical patterns and patient representation learning. In \cite{bengio2013representation}, the authors develop a deep neural network composed of a stack of denoising autoencoders to extract relevant patterns from Electronic Health Records (EHRs). 
Meanwhile, to handle vanishing exploding gradient problems, different variants of RNN have been proposed. 
For example, Long-Short Term Memory (LSTM) \cite{hochreiter1997long}, one of such popular variants which can handle long-term event dependencies by utilizing a gated architecture, has recently been applied to health informatics \cite{che2015deep,che2017rnn,che2018recurrent} with promising results.

Despite such successes, discovering hidden patterns of the sequential data is still an open challenge since it requires intelligent segmentation and clustering of the time series data.  
For example, the time lapse between successive elements in patient records can vary from days to months, which may lead to suboptimal performance for the traditional LSTM models.
In addition, it is also difficult to interpret their impressive performance, especially when the data is high-dimensional, which in turn 
%raises concerns of the lack of interpretability and 
limits the ability to design better architectures. 
To address these challenges, three research questions have been raised as follows: 

\begin{itemize}

\item (1) RQ1: ``Can we predict diseases status based on individual records?"

\item (2) RQ2: ``How to build an appropriate learning model that can deal with the irregular data collection times, and learn hidden patterns from time series features?"

\item (3) RQ3: ``How does one interpret such a predictive model?"
 
\end{itemize}

In order to tackle those research questions, in this paper, we take the obesity disease as a use case and solve an obesity status improvement prediction task using activity data collected via a mobile phone application named FeatForward. We recruit 275 participants and collect activity data using this phone application for 6 months.
The application collect useful information such as previous diagnoses, blood test results, and activity levels (e.g., step counts).

Then, based on these collected data, we develop a Recurrent Neural Network (RNN) based time-aware learning model that performs obesity status improvement prediction.
In our model, we introduce a day-week-month variable to deal with irregular data collection times, and a person's current and past obesity status are used to predict his/her obesity status in future. 
In addition, we interpret the behavior of our learning model by analyzing the expected responses of the hidden state units given certain inputs.
%between a hidden state unit and the input 
%are measured using the expected updates or responses of the hidden state units given the input.
Finally, we evaluate the performance of our proposed model using a real-world dataset. The experimental results show that our proposed method can capture the underlying patterns in users' time series with irregular data collection time instances, and achieve an accuracy of 77-86\% for the obesity status improvement prediction. 
%This in turn provide evidence that data collected from commercial mobile health devices can yield useful information on the relationship between steps and health outcomes.

Our major contributions are:

\begin{itemize}
\item In order to conduct a general study, we use a mobile application named FeatForward that developed by our team to collect data from 275 participants within a 6 months period and build a RNN based time-aware predictive model to forecast individuals' obesity status.

\item In order to solve the irregularity in data collection time instances, we also directly incorporate the day-week-month effect to improve the prediction performance.

\item We design and implement an analytical framework to understand how the features contribute to our obesity status improvement predictive model.

\item We provide extensive experimental results using real-world data to show that our mechanism is accurate and widely applicable.

\end{itemize}

The rest of the paper is organized as follows. Section II discusses related work. Section III provides brief descriptions of our data collection program. Section IV  describes the proposed method in detail. Section V presents the evaluation results of our approach with real-world data. Section VI concludes the paper and highlights our future directions.

%\input{related.tex}

%\vspace{-0.06in}
\section{Related Work}
%\vspace{-0.06in}
\label{sec:related}

In this section, we briefly review existing works in healthcare, which are closely related to our proposed method in this paper from two areas. The first one is recent works on clinical data mining and exploiting deep learning methods in the healthcare domain. 
The other is the existing explainable deep learning approaches that have utilized visualization to help understand machine learning models.

\subsection{Clinical Data Mining}
%This sub-section presents several healthcare applications, services, and systems that are supported by data analytics in electronic health records (EMR) and sensor data. 

\subsubsection{Conventional Machine Learning on Health Data} \
Clinical Data Mining (CDM) is the application of data mining techniques using clinical data \cite{iavindrasana2009clinical}, to extract relevant knowledge and make clinical decisions \cite{epstein2009clinical}. 
In \cite{aronson2010overview} Carreiro et al. considered the database as a social network, and try to extract relevant information from related communities. 
Marwa et al. in \cite{elamin2015predicting} have also presented an algorithm to generate prediction models based on the information gathered on patient's first physician visit. 
In addition, previous studies (\cite{ben2015next,chow2016sad,huang2016assessing}) have used individual physical details including behavioral, sleeping, voice acoustic, and social patterns to estimate a person's mood.
While their mechanisms are beneficial, they ignore the long-term dependencies among collected data records, which can be learnt using deep learning, a more robust learning tool.

\subsubsection{Deep Learning on Health Data} \
As deep learning has achieved great success recently, researchers have begun attempting to apply neural network based methods to clinical temporal health data \cite{lasko2013computational,razavian2015temporal}.
For example, the authors in \cite{cheng2016risk} proposed an adjustable temporal fusion scheme using CNN-extracted features to process EHR for risk predictions. 
Meanwhile, Recurrent Neural Networks (RNNs) also can be used for disease prediction with times series data in healthcare domain.
For example, RETAIN and GRAM in \cite{choi2016retain,choi2017gram} are two state-of-the-art models utilizing RNNs for future disease predictions. 
In \cite{pham2016deepcare}, Pham et al. introduced an end-to-end deep network called ``Deep-Care", which have used LSTM for predicting future admission of a patient, and also addressed the time irregularities between consecutive entries. 
In addition. Suhara et al. in \cite{suhara2017deepmood} developed a deep learning based approach that forecasts severely depressive mood based on individual's historical moods, behavioral type, and medication records.
While those models achieve good prediction performance, they do not exploit mobility or activity patterns for diseases monitoring and lack model interpretations.

\subsection{Models Interpretation}

In the field of visualization, most of the existing explainable deep learning approaches mainly focus on understanding and analyzing model predictions or the training process offline after the model training is completed. 
Recent work (\cite{karpathy2015visualizing,sha2017interpretable}) has exhibited the effectiveness of visual analytics in understanding, diagnosing and presenting neural networks. In \cite{liu2017towards}, Liu et al. treated deep CNN as a directed acyclic graph and built an interactive visual analytics system to analyze CNN models. Rauber et al. in \cite{rauber2017visualizing} applied dimensionality reduction to visualize learned representations, as well as the relationships among artificial neurons and provided insightful visual feedback of artificial neural networks. 
Meanwhile, in \cite{greff2017lstm} Greff et al. conducted a comprehensive study of LSTM components and Chung et al. in \cite{chung2014empirical} have evaluated GRU compared to LSTMs. 
While visualizations have achieved considerable success on deep learning models, they only provide an overall analysis without exploring hidden states in detail and are not scalable when hidden state dimensions have increased.

%\input{data_collection.tex}

%\vspace{-0.06in}
\section{Data Collection}
%\vspace{-0.06in}
\label{sec:dc}

%With the advancement in sensor technology and miniaturization of sensor devices, various types of tiny, energy-efficient and low-cost sensors are expected to be widely used for improving healthcare \cite{alemdar2010wireless,caytiles2014study,filipe2015wireless}. 

%\vspace{-0.1in}
\begin{figure}[!htbp]
\centering \includegraphics[height=1.6in,width=2.8in]{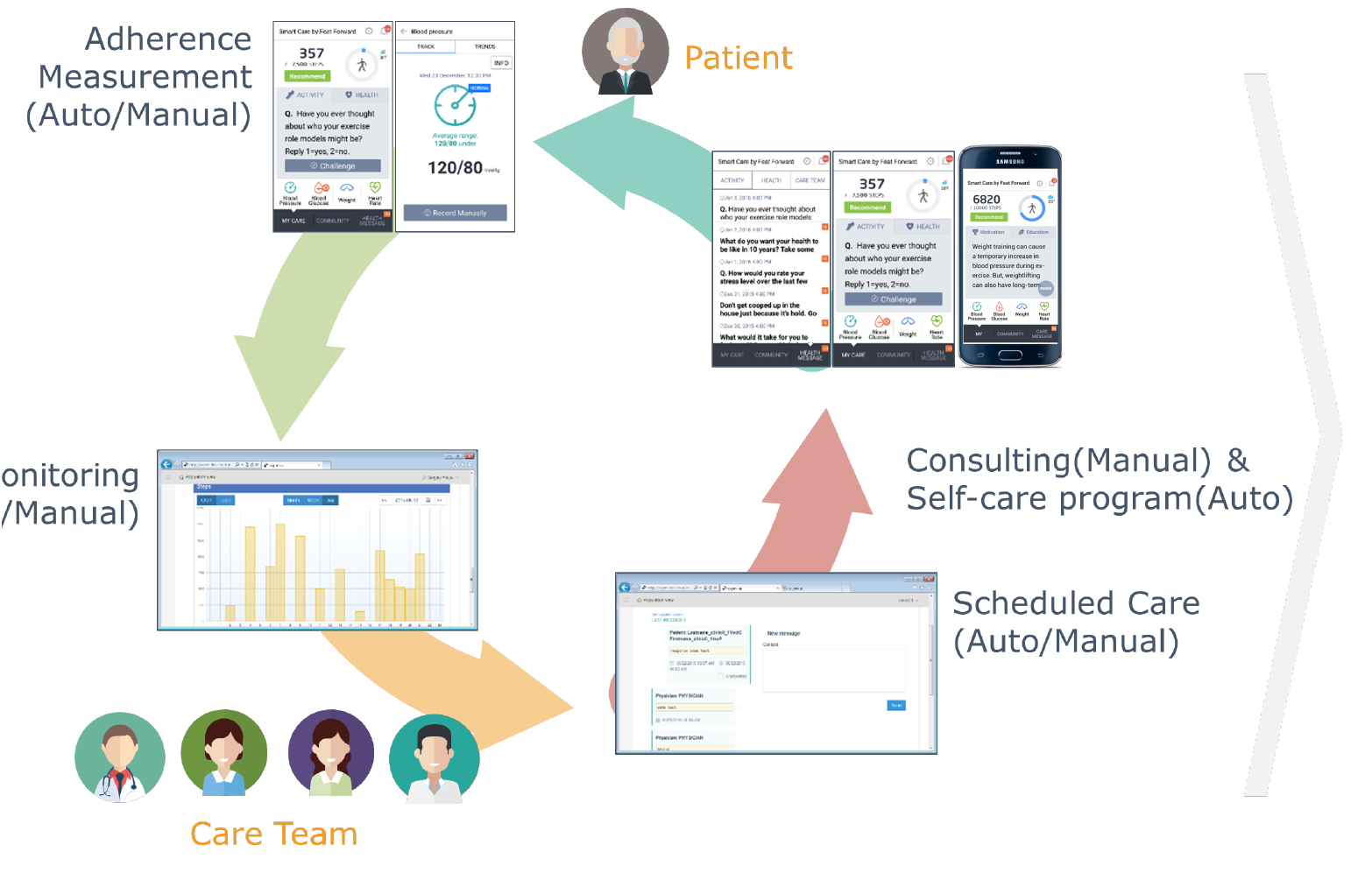}
\vspace{-0.1in}
\caption{\small FeatForward Study Overview}
\label{fig:app}
\vspace{-0.25in}
\end{figure}

In our work, we use a smartphone application called FeatForward that was developed by our team to collect data from 275 participants for 6 months. 49.8\% of the participants are over 50 years old and 50.2\% are below 50. Such data allows us to infer three highly prevalent but often undiagnosed conditions including diabetes, hypertension, and obesity that participants may have.
The application provides an intuitive interface (Fig~\ref{fig:app}) for users to record their data including gender, age, weight, blood pressure, etc. at three milestones (enrollment, midpoint, closeout).
It also allows participants to voluntarily record their activity information (step counts) at different time slots (e.g., every 20 minutes or hourly) every day. The data can be divided into two 3-months periods, the $1^{st}$ one is from enrollment to midpoint while the $2^{nd}$ one is from midpoint to the end of the study. The application was designed such that participants will be shown the trends of their health status at the end of that 3 months and be encouraged to increase their activity levels if they cannot achieve the health related goals (expressed in terms of clinically relevant threshold (CRT) which is a measure of the percentage reduction in their BMI values). 

\begin{table*}
\centering
\caption{\small Collected Data Features}
\vspace{-0.1in}
\begin{tabular}{|p{2.4cm}|p{5.5cm}|p{3.5cm}|p{5.0cm}|}
\hline\hline
\footnotesize \centering Classification & \footnotesize \centering Detailed & \footnotesize \centering Collection & \footnotesize Notes \\ [0.1ex]
\hline
\footnotesize \centering Personal Data & \footnotesize \centering gender, age, height, disease & \footnotesize \centering manual input & \footnotesize occurs at registration  \\
\hline
\footnotesize \centering Biometric & \footnotesize \centering step count & \footnotesize \centering once per hour & \footnotesize obtained through S-Health \\
\cline{2-4}
\footnotesize \centering Data & \footnotesize \centering blood pressure, blood sugar, weight, heart rate & \footnotesize \centering manual input & \footnotesize through the S-Health GUI input window \\
\hline
\footnotesize \centering Messaging & \footnotesize \centering activity/education & \footnotesize \centering whenever event occurs & \footnotesize automatic program transmission \\
\cline{2-4}
\footnotesize \centering History & \footnotesize \centering communication & \footnotesize \centering whenever event occurs & \footnotesize foctor-patient messaging  \\
\hline
\footnotesize \centering Environment & \footnotesize \centering weather information & \footnotesize \centering & \footnotesize weather forecast \\
\hline
\end{tabular}
\label{tab:collectfeatures}
\vspace{-0.2in}
\end{table*}

In our pilot study, we required all participants to wear the provided smartwatch during all hours except while they sleep for accurate and comprehensive data collection.
%to use the phone as their primary phone for comprehensive data collection 
The recorded information is shown in Table~\ref{tab:collectfeatures}, and  can be visualized in an aggregated manner so that participants can observe the general trends of their health status.
Thus, through FeatForward, participants will be able to assess the effects of physical activities on changes in their health status which may in turn encourage them to be more physically active.

%\input{method.tex}

%\vspace{-0.06in}
\section{Methodology}
%\vspace{-0.06in}
\label{sec:method}
In this section, we consider the obesity disease as a use case and utilize the collected data from the FeatForwad pilot study (described in section III) to generate clinical learning models to predict obesity status improvement and demonstrate the usefulness of using activity data collected via wearables for predicting health status. 

\subsection{Preliminaries}

Different from other application domains (e.g., image and speech analysis), the problems in healthcare are more complicated. For example, the diseases are highly heterogeneous which make it hard for physicians to understand their causes and how they progress completely. 
Thus, in this sub-section, we first discuss the three research questions we raised in section I that helps to define our design requirements.

\subsubsection{RQ1: Can we forecast obesity disease status based on individual records?}

The question includes identifying specific measurements which can contribute towards improving prediction performance.
In \cite{tryon2013activity}, Tryon has noted that step count information is a preferred metric for quantifying physical activity. Additional risk factors which impact prediction results that need to be considered include users' demographics and health histories e.g., his/her height, age, and fitness level.

\subsubsection{RQ2: How to generate an appropriate model that can deal with irregular  observation times?}

In the pilot study, participants record their activity information at different time slots each day, and how their BMIs vary at the 3 milestones. Such raw data collected via the FeatForward application
can be broken down into segments of time series data and mined for predicting obesity status improvement.
For example, the obesity status prediction can be described as a $(m, n)$ task, where a feature vector $x^P_d$ can be extracted from data recorded in day $d$ of a participant $P$. $x^P_d=\{x^P_{d_1}, x^P_{d_2}, \dots, x^P_{d_m}\}$, where $x^P_{d_m}$ is a sequence of feature vectors of a participant $P$ collected at time slot $m$ of day $d$. 
The change of a participant's BMI value (a measure of obesity status improvement) will be predicted based on his/her histories in the previous $n$ days $\{x^P_1, x^P_2, \dots, x^P_n\}$.
%Since there is an abundance of machine learning models, including neural networks that have been applied in the healthcare domain due to their high performance on time series inputs, so in our work, 
In this work, we proposed a RNN based time-aware model to learn long-term dependencies among time series features to perform such a prediction.

\subsubsection{RQ3: How to interpret such a learning model?}

While recent developments in the health informatics community have seen widespread adoption of modern machine learning methods, relatively little attention has been paid to understand the properties of its representations and predictions. 
Thus, in our work we try to interpret the changes of individual hidden states based on different inputs.
%Thus, to alleviate those issues, we have proposed our methodology to create useful training models for obesity status forecasting, which can be easily understood.

\vspace{-0.15in}
\begin{figure}[!htbp]
\centering \includegraphics[height=0.85in,width=3.5in]{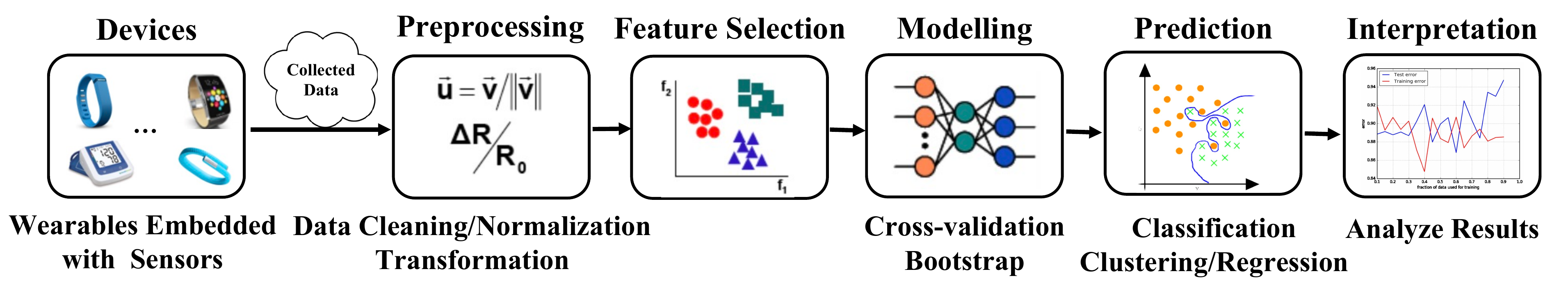}
\vspace{-0.2in}
\caption{\small Workflow of Our Method}
\label{fig:overview}
\vspace{-0.18in}
\end{figure}

\subsection{Methodology Overview}

Before participants' records are input into various models for analysis, data needs to go through several steps of processing, as shown in Fig~\ref{fig:overview}.
%Firstly, participants' histories need to be recorded, accessed and acquired. 
First, data cleaning and transformation need to be performed since participants may not record measurements in all time slots and measurements may be either discrete or continuous values.
Next, feature selection and extraction are highly desirable to produce learning models with high prediction accuracy.
Finally, the learnt model need to be interpreted to uncover the impact of the hidden units on the prediction results. 

\subsection{Detailed Design}
In this sub-section, several key steps for processing data, namely, data cleaning, data transformation, feature extraction and data analytics/modeling are described in detail respectively.

\subsubsection{Data Cleaning \& Normalization}
Once the data has been collected, it needs to be imputed and normalized.
These processes are expected to improve the data quality which helps to improve the accuracy of the learning models.
Typically, missing data may be caused by either participants forgetting to wear the sensor devices or fail to record their data through the mobile application.
In our work, we replace any missing value with the average value obtained from non-missing entry values of that feature.

\subsubsection{Data Transformation}
Since participants record their activity information (step counts) at different time slots in each day, it is challenging to perform data analytics with such irregular observation times.
To tackle this problem, in our work, we introduce a day-week-month variable to organize participants' data into different non-overlapping time ``windows" with each window being $k$ days.
% and the regularly reorganized information is then used for further modelling and analysis. 
While transforming variable data into fixed-length time series allows us to employ some efficient methods directly, we need to be aware of the side effects associated with such segmentation. For instance, unwise segmentation may result in the sparsity and missing data problems since there could be no observations for some features in some time windows. In addition, by dividing longitudinal data into ``windows", the model may be less sensitive to capturing long-term feature patterns. 

\subsubsection{Feature Selection \& Extraction}

Since no one knows precisely which features (participants' characteristics or activity information) are more critical for obesity status prediction, we conduct feature selection operations to assess their usefulness in constructing the learning models.  
Cross-sectional studies \cite{bassett2017step} have shown that the daily step count is inversely related to body mass index (BMI), hypertension, and diabetes and hence will be considered.
% which can be used as an overall measurement of physical activity for collected data analytic tasks.

While daily average step feature plays an essential role in referring to obesity risk, this metric does not tell us the frequency, intensity, or duration of a person's physical activity. Thus, we also include additional related features: (i) intensity: maximum or minimum steps per day/week; (ii) frequency: number of days where the daily step counts exceed a threshold (set to twice the average lowest daily step counts); (iii) duration of physical activity:  a number of days that the participants will walk more steps than the average step count when compared with other participants. 
In addition, we also consider a patient's demographics since the steps that a person takes vary based on his/her height, age, and fitness level. 
For example, frail, elderly individuals tend to take slower and fewer steps while younger individuals often take more steps.
\begin{table}
\centering
\caption{\small Extracted Features}
%\vspace{-0.1in}
\begin{tabular}{|p{1.4cm}|p{1.0cm}|p{5.2cm}|}
\hline\hline
\footnotesize \centering Features & \footnotesize \centering Type & \footnotesize Description \\ [0.1ex]
\hline
\footnotesize \centering & \footnotesize \centering Time Slot & \footnotesize step count at every time slot \\
\cline{2-3}
\footnotesize \centering & \footnotesize \centering & \footnotesize day: average steps, maximum/minimum steps \\
\cline{3-3}
\footnotesize \centering & \footnotesize \centering & \footnotesize week: maximum/minimum steps \\
\cline{3-3}
\footnotesize \centering Step & \footnotesize \centering & \footnotesize total steps \\
\cline{3-3}
\footnotesize \centering Information & \footnotesize \centering Extracted & \footnotesize how frequently does the participant walk \\
\cline{3-3}
\footnotesize \centering & \footnotesize \centering Features & \footnotesize \# of days that has larger steps than the average steps (per day/week) of a specific participant \\
\cline{3-3}
\footnotesize \centering & \footnotesize \centering & \footnotesize \# of days that has larger steps than the average steps (per day/week) of all participants \\
\hline
\footnotesize \centering Demographic & \footnotesize \centering & \footnotesize gender, age, marital, adult in household, highest degree, hispanic or latino, race, occupation \\
\hline
\end{tabular}
\label{tab:selectfeatures}
\vspace{-0.1in}
\end{table}
Finally, all the features extracted from collected data and used in our work are shown in Table~\ref{tab:selectfeatures}.

\subsubsection{Model Construction \& Interpretation}
After data cleaning, transformation, and feature extraction, we then build learning models for obesity status improvement prediction, where we pose our predictive problem as a sequence classification task.
%For example, given multivariate time series $X^P=\{x^P_1, \dots, x^P_n\}$ of tracked behavioral data for $n$ days for a participant $P$, we estimate the conditional probability $p(Y|X^P)$ of the target $Y$ (e.g., whether the change of participant's future BMI is above a certain threshold).
%\vspace{-0.05in}
\begin{figure}[!htbp]
\centering \includegraphics[height=1.8in,width=3.6in]{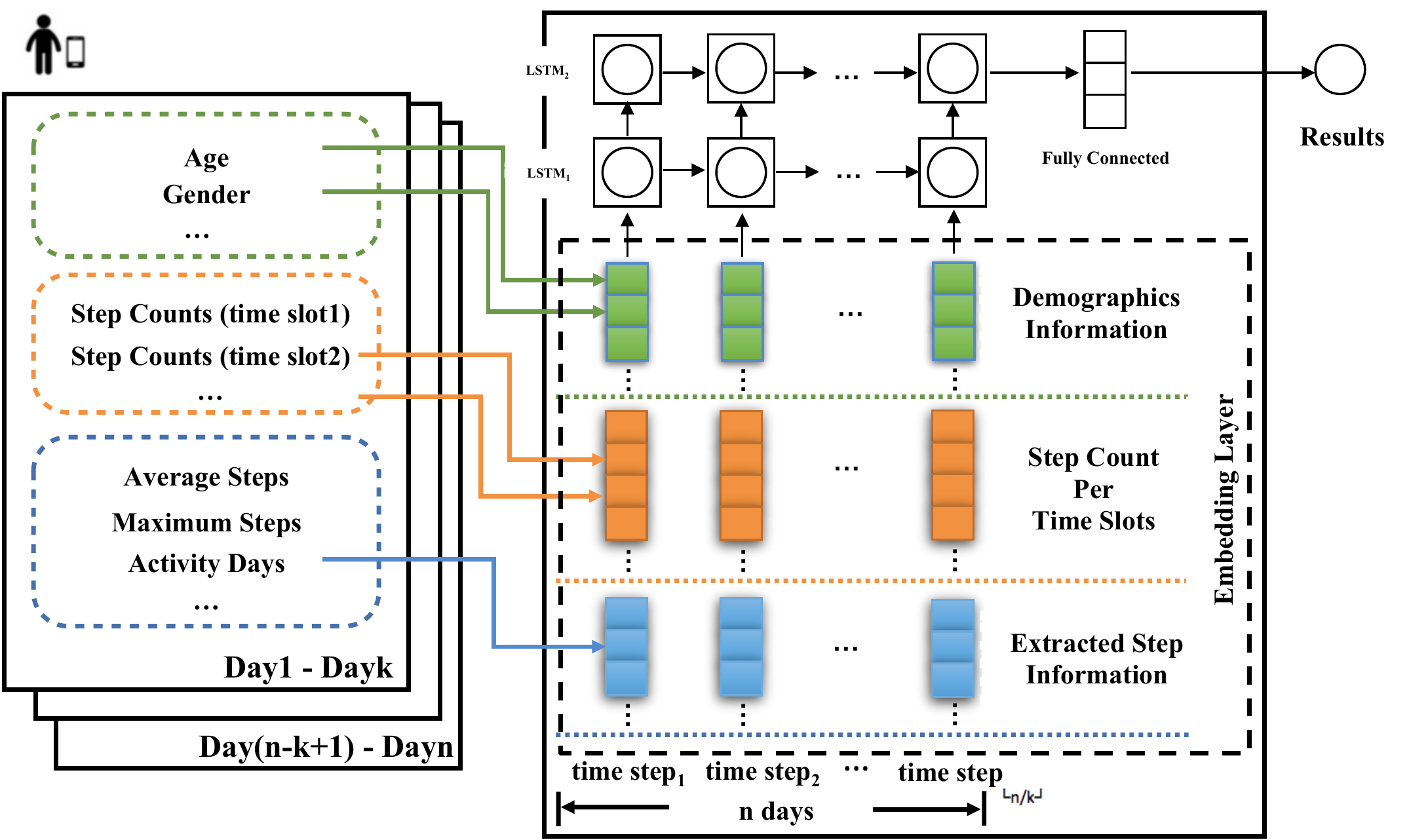}
\vspace{-0.2in}
\caption{\small Network Architecture of Our Method}
\label{fig:model-construction}
\vspace{-0.2in}
\end{figure}
To quantify the effectiveness of the neural network models, we first develop a set of baseline classifiers using traditional machine learning approaches such as Logistic Regression (LR) and Random Forest (RF) so that we can later compare them with those obtained using neural network models. 
%and learn their feature representations.

{\bf (1) Baseline Classifiers}

The reason why we select LR and RF as our baseline classifiers is because: (i) the L2-regularizer in LR classifier 
%penalizes large weights and make the classifier 
is more robust to limited training examples;
(ii) the RF classifier is a robust and ensemble-based machine learning method, which is a commonly used approach in the healthcare domain.
However, those commonly used classification approaches assume all features are independent and do not consider the impact of their time correlations.

{\bf (2) Deep Learning Models} 

In order to use individual histories as time series data, we need a technique that is capable of incorporating dependencies from previous states.   
In our work, we consider a Long Short-term Memory (LSTM) network, which captures correlations across different behavioral sequences over time.

%We first introduce a time window $k$ to distinguish the same activities during the same period of time on different $n$ days.
First, we segment the collected time series data into different segments, each of which lasts for $k$ days and generate corresponding features from the collected records in each segment, as shown in Table II.
For instance, $k=7$ means the data collected every week is used to generate useful features, e.g., the average step counts per time slot per week, the average daily step counts, the maximum weekly step counts, etc.

Because we have different type of features with distinct semantics, we also introduce an embedding layer to convert those raw data into a dense vectorial representation.
The outputs of the embedding layer will be fed into a 2-layer LSTM model, where the first layer is used to extract features from the vectorial representation and the $2^{nd}$ layer is used to learn high level abstraction representations. The output of the $2^{nd}$ layer is fed to a fully connected layer to perform prediction.
%Finally, the LSTM layer will propagate the historical information along with the inputs of the next step until it reaches the fully connected layer.

The network architecture of our method is illustrated in Fig~\ref{fig:model-construction}, where all inputs grouped every $k$ days are fed into the model and prediction results are generated as the output $Y_{\lfloor n/k \rfloor}$ at the final time step $\lfloor n/k \rfloor$.

{\bf (3) Learning Models Interpretation} 

\vspace{-0.1in}
\begin{figure}[!htbp]
\centering \includegraphics[height=1.3in,width=1.6in]{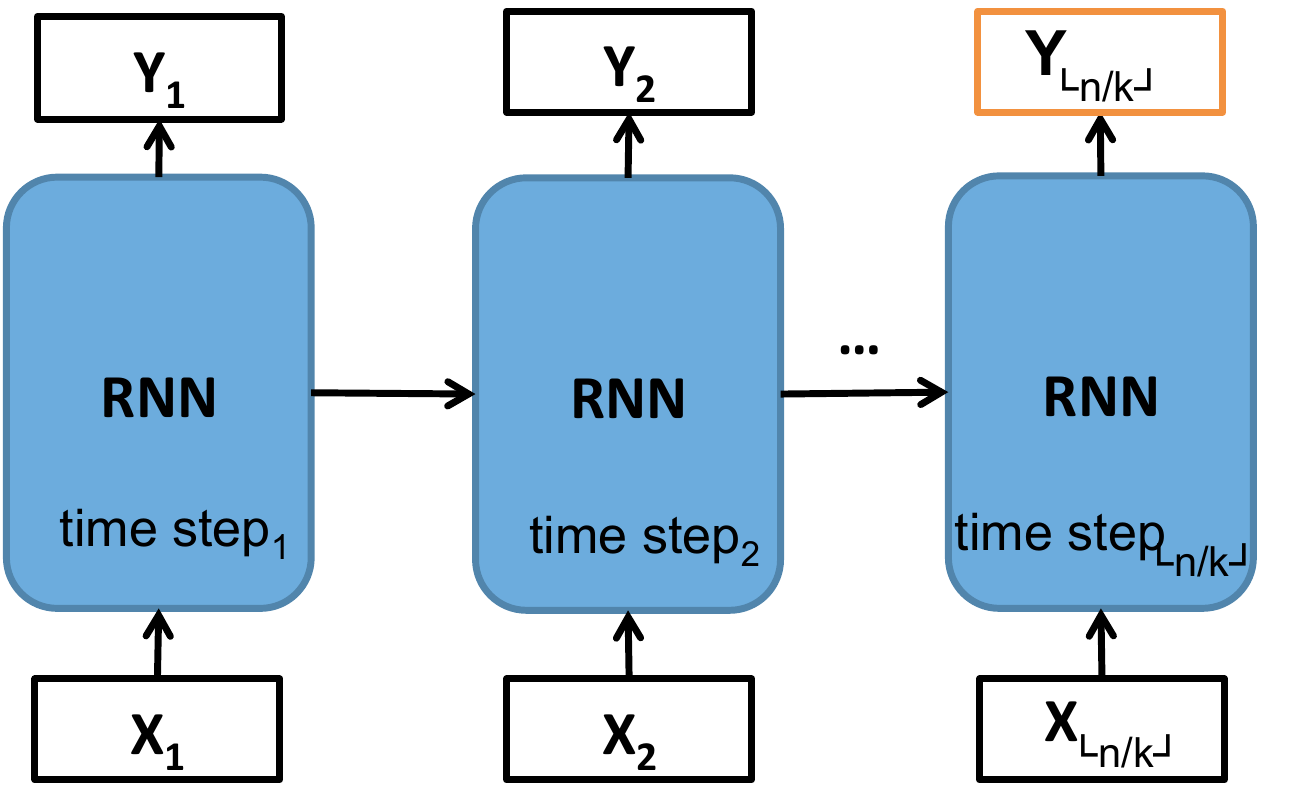}
\vspace{-0.05in}
\caption{\small Input-output Scheme}
\label{fig:input-output-scheme}
\vspace{-0.1in}
\end{figure}

Inspired by the idea of using interpretable representations \cite{ming2017understanding} to explain functions of network components, we use a similar method to interpret the learned representation of our proposed model.
Our model has a sequence-to-sequence input-output scheme, as shown in Fig~\ref{fig:input-output-scheme}.
This formulation takes an input sequence $X_t=\{X^1_t, X^2_t, \dots, X^{|X_t|}_t\}$ and generates an output $Y_t$ at every time step $t$, which can be further processed as the input of the next time step or directly used for classification tasks.
In our work, we only use the output at the last time step $\lfloor n/k \rfloor$ of the whole sequential inputs $\{X_1, X_2, \dots, X_{\lfloor n/k \rfloor} \}$ for the prediction task.

%To confirm the learned representation, we look up the representative input for each time step $t$, where we calculate the model's expected responses based on the update of hidden states. The reason of this specification is that 
At each time step $t$, the model takes an input $X^i_t$, and updates the hidden state using:

%\vspace{-0.1in}
\begin{equation}\label{eq1}
\begin{array}{lcr}
h^i_{t} = f(Wh_{t-1}+VX^i_t)
\end{array}
\end{equation}
%\vspace{-0.2in}

where $W$ and $V$ are weight matrices and $f$ is a nonlinear activation function.
After the updates, $h^i_t$ is considered to capture the long-term memory and used to compute the intermediate or final output.
Although $h^i_t$ is calculated by a non-linear transformation of the previous state $h_{t-1}$ and the input $X^i_t$, $\triangle h^i_t=h^i_t-h_{t-1}$ is deterministic with regards to the input $X^i_t$ when the previous history $h_{t-1}$ is given, and can be used to reflect the degree a hidden state in the model is influenced by the input.
Thus, based on $\triangle h^i_t$ the model's response to the input $X_t$ can be computed as follows:

%\vspace{-0.1in}
\begin{equation}\label{eq1}
\begin{array}{lcr}
s(X_t) = E(\triangle h_t| X_t) = \frac{\sum_{i=1}^{|X_t|} \abs{\triangle h^i_t}}{\sum_{i=1}^{|X_t|} 1}
\end{array}
\end{equation}
%\vspace{-0.13in}

where $s(X_t)$ is the average absolute expected responses of $|X_t|$ input sequences from all participants.
Note that with the $tanh$ activation function, the response can have either positive or negative value, where a larger absolute value of response indicates that the input is more salient to the corresponding hidden state.

%\input{evaluation.tex}

%\vspace{-0.06in}
\section{Performance Evaluation}
%\vspace{-0.06in}
\label{sec:eval}
%\vspace{-0.1in}

In this section, we analyze the collected dataset to confirm that it contains useful information to infer participants' obesity status improvement levels as well as verify that the model we suggest can make useful predictions. We first describe our experimental settings and the data we use. Then, we provide the preliminary results in dealing with the research questions that described in section I.

\subsection{Experimental Setup}
To evaluate the performance of our scheme, we conduct experimental evaluations on real-world data, which is collected by the FeatForward application described in Section III. It tracks 275 participants for a 6-month period which can be divided into two 3-months periods, namely from enrollment to midpoint ($1^{st}$ 3 months) and from midpoint to end of the study ($2^{nd}$ 3 months). The measured metrics consist of an individual's time series of daily step counts (measured using 20 minutes time windows), his/her medical records and demographics (as shown in Table II). One can use statistics from different time windows to derive learning models. For example, we can conduct training using measurements collected with different day time windows ($k=$ one day, one week, one month) and daily time slots (i.e., $m=4, 6, 12$). 
Since some participants dropped out after the midpoint or they do not record enough step counts information, so after data cleaning, we obtain dataset $D_1$, which contains 323 instances. Each instance contains historical measurements of a participant for 3 months. Next, we conducted 10-fold cross validation, where the cross validation splits the datasets into training (80\%), validation (10\%) and testing sets (10\%). Our training model is used to predict whether a participant's BMI change exceeds the clinically relevant threshold (CRT) at the end of 3 months.
For our training, we apply a dropout mechanism and conduct regularization to overcome any possible overfitting problem.

In addition, since in the collected datasets, participants share their BMI measurements,  we can use such measurements to label each participant as positive or negative instance depending on whether their BMI change percentage exceeds the CRT. In our case, we set CRT to 5\% since past studies, e.g, \cite{nhi} have shown that 
it is reasonable to achieve an average weight loss of 5-10\% during 3-6 months.
%It also mentions that ``weight loss at the rate of 1 to 2 lb/week (calorie deficit of 500 to 1000 kcal/day) commonly occurs for up to six months, at which point weight loss begins to plateau unless a more restrictive regimen is implemented".
%Thus, in our experiments we assume that participants' weights will adjust linearly within six months.
%Since our datasets contain two time durations, where each lasts three months, so we set a clinically relevant threshold ($CRT$) as 5\% and label participants into two groups based on their BMI changes as follows: (i) we assign positive labels to the participants' data instances if the difference of their BMI values are larger than $CRT$ during every time duration (enrollment to midpoint or midpoint to closeout); (ii) Otherwise, we label the data instances as negative.
With such labeling method, we obtain 53\% positive instances and 47\% negative instances.

For our obesity status improvement learning model, we use 25 hidden units for our LSTM and apply a dropout of 0.5. We train our model for 150 epochs, where each epoch is defined as the process of feeding the whole training set to a model. 
All our experiments are conducted on Mac Pro with an Intel Core i7 processor running at 2.5GHz, 16GB memory and an external GTX 1080.

\subsection{Performance Evaluation}

\subsubsection{Importance of Step Counts in BMI changes}

First, we conducted Exp1 using the $D_1$ dataset to measure the relationship between step counts and BMI changes.
 
Exp1: In this experiment, we first compute the daily average step counts $DAS^P$ for every participant $P$. Then, we use an IBM SPSS 20 tool to examine the association between participants' daily average step and changes in obesity related indicators (e.g., BMI) using the linear mixed-effects models \cite{vemuri2009mri}, a preferred method for evaluating the longitudinal effects over time.
We also compute the daily average step counts for all participants, denoted as $DAS^{all}$ and compute the probability that a participant's BMI will drop after 3 months based on the comparison between his/her daily average step counts with $DAS^{all}$.

\begin{table}
\centering
\caption{\small Linear Mixed-effect Model}
\vspace{-0.1in}
\begin{tabular}{|p{1.2cm}|p{1.9cm}|p{2.1cm}|p{1.0cm}|p{0.5cm}|}
\hline\hline
\footnotesize \centering Source & \footnotesize \centering Numerator (df) & \footnotesize \centering Denominator (df) & \footnotesize \centering F & \footnotesize sig \\ [0.1ex]
\hline
\footnotesize \centering Intercept & \footnotesize \centering 1 & \footnotesize \centering 105 & \footnotesize \centering 2533.879 & \footnotesize .000 \\
\hline
\footnotesize \centering Step & \footnotesize \centering 1 & \footnotesize \centering 105 & \footnotesize \centering 5.492 & \footnotesize .021 \\
\hline
\end{tabular}
\begin{tablenotes}
\vspace{-0.03in}
\item{\footnotesize a. Dependent Variable: BMI};
\end{tablenotes}
\label{tab:linear-mix-effect}
\vspace{-0.2in}
\end{table}

The results of the linear mixed-effect model are shown in Table~\ref{tab:linear-mix-effect}. From the results we can see that the individual daily average step count is highly predictive of BMI changes.
%Such conclusion can also be confirmed by the results of the statistical analysis, where we find that 
From our analysis of the participants' records, we observe that any participant $P$ whose daily average step $DAS^P$ is larger than $DAS^{all}$ has a high probability (73\% on the average) to have a drop in his/her BMI value after 3 months.

\subsubsection{RQ1: Can we forecast obesity health status improvement? How to select essential features?}

In this subsection, we conduct Exp2 to evaluate the forecasting performance of traditional models. Such performance results are used as baselines for comparison with deep-learning based model in subsequent subsection.

Exp2: We first use the traditional classifiers LR and RF (described in section IV) to generate the learning models using different combinations of extracted features (Table II) so that we can compare their performances: (i) we only use daily average step feature to build the models; (ii) instead of using the daily average step count feature, we use step counts measured using $m$ different time slots; (iii) we use both time slot features and demographics to train the learning models.
In this experiment, we use dataset $D_1$ with different $m$ values ($m=4, 6, 12$). 

\begin{table}
\centering
\caption{\small Impact of Different Selected Features}
\vspace{-0.1in}
\begin{tabular}{|p{3.0cm}|p{0.6cm}|p{0.7cm}|p{1.2cm}|p{1.2cm}|}
\hline\hline
\footnotesize \centering Selected Features & \footnotesize \centering Time Slot & \footnotesize \centering Feature Size & \footnotesize \centering Accuracy (LR) & \footnotesize Accuracy (RF) \\ [0.1ex]
\hline
\footnotesize \centering Time Slots & \footnotesize \centering 4 & \footnotesize \centering 4 & \footnotesize \centering 54.4$\%$ & \footnotesize 53.1$\%$ \\
\hline
\footnotesize \centering Time Slot+Demographic & \footnotesize \centering 4 & \footnotesize \centering 12 & \footnotesize \centering 55.9$\%$ & \footnotesize 54.7$\%$ \\
\hline
\footnotesize \centering All Features & \footnotesize \centering 4 & \footnotesize \centering 23 & \footnotesize \centering 59.0$\%$ & \footnotesize 63.1$\%$ \\
\hline
\footnotesize \centering Time Slots & \footnotesize \centering 6 & \footnotesize \centering 6 & \footnotesize \centering 54.2$\%$ & \footnotesize 56.5$\%$ \\
\hline
\footnotesize \centering Time Slot+Demographic & \footnotesize \centering 6 & \footnotesize \centering 14 & \footnotesize \centering 56.2$\%$ & \footnotesize 57.8$\%$ \\
\hline
\footnotesize \centering All Features & \footnotesize \centering 6 & \footnotesize \centering 25 & \footnotesize \centering 60.6$\%$ & \footnotesize 63.7$\%$ \\
\hline
\footnotesize \centering Time Slots & \footnotesize \centering 12 & \footnotesize \centering 12 & \footnotesize \centering 53.8$\%$ & \footnotesize 53.2$\%$ \\
\hline
\footnotesize \centering Time Slot+Demographic & \footnotesize \centering 12 & \footnotesize \centering 20 & \footnotesize \centering 55.1$\%$ & \footnotesize 53.8$\%$ \\
\hline
\footnotesize \centering All Features & \footnotesize \centering 12 & \footnotesize \centering 31 & \footnotesize \centering 56.9$\%$ & \footnotesize 60.2$\%$ \\
\hline
\end{tabular}
\label{tab:diffeatures}
\vspace{-0.1in}
\end{table}

The results are shown in Table~\ref{tab:diffeatures}. From the results, we can observe that each of the two classifiers (LR or RF) achieves significantly higher accuracy than random guessing (e.g., 0.5). It is better to use all features to generate learning models since the obesity status improvement depends not only on the number of step counts but also on the frequent, intensity, duration of a participant's physical activity bouts as well as his/her demographics.
%For example, in the logistic regression model, an important feature or a more unique demographic will be associated with a higher coefficient value, which indicates its marginal contribution to the predicted risk. 
In addition, the performance of RF is better than LR (by 3\% on average) when all features are used. It is expected since RF is less likely to overfit and it also learns better the correlations among different features.
% not only the problem of over-fitting can be overcome to a large extent by using  RF, but also the RF can learn the strong interactions among features.
Moreover, we find that the performance of all models varies when $m$ increases from 4 to 12, with the best performance being achieved when $m=6$. This can be explained as follows: more noises in the training data will be observed
with a larger number of time slots (e.g., 12) since more time slots will have zero measurements.  With fewer number of time slots (e.g., 4), the observed step counts for different time slots may not differ much from one another and hence affects the prediction results. 

\subsubsection{RQ2: Which model performs best in obesity status improvement predictions?}

We conduct two experiments (Exp3.1 \& Exp3.2) to examine how the selection of window size $k$ affects our results. For example, $k=30$ means that the input of a predictive model is trained based on the data extracted from every month.

Exp3.1: Since those traditional classification approaches (Exp2) assume all features are independent and do not consider the time correlation of these features, we use a single layer LSTM architecture in this experiment so that we can see if such learning model can perform better than traditional methods. 
We also set $n=90$, $k=3, 7, 30$ and assign $\lfloor n/k \rfloor$ (i.e., 30, 12, 3) time steps for such LSTM model, where the input of each node is the related feature vector generated based on $k$ days. 

Exp3.2: Instead of only using a single layer LSTM as in Exp3, we use other learning models, which include a 2-layer CNN model and multi-layer LSTMs.

\begin{table}
\centering
\caption{\small Impact of Different Time Window Sizes}
\vspace{-0.1in}
\begin{tabular}{|p{1.05cm}|p{0.4cm}|p{0.7cm}|p{0.95cm}|p{0.55cm}|p{1.35cm}|p{0.9cm}|}
\hline\hline
\footnotesize \centering Model & \footnotesize \centering Time Slot & \footnotesize \centering Feature Size & \footnotesize \centering Window Size & \footnotesize \centering Step Nodes & \footnotesize \centering Accuracy (Validation) & \footnotesize Accuracy (Testing) \\ [0.1ex]
\hline
\footnotesize \centering LSTM (1 layer) & \footnotesize \centering 6 & \footnotesize \centering 25 & \footnotesize \centering 3 days & \footnotesize \centering 30 &\footnotesize \centering 61.3$\%$ & \footnotesize 60.9$\%$ \\
\hline
\footnotesize \centering LSTM (1 layer) & \footnotesize \centering 6 & \footnotesize \centering 25 & \footnotesize \centering 1 week & \footnotesize \centering 12 & \footnotesize \centering 69.4$\%$ & \footnotesize 68.7$\%$ \\
\hline
\footnotesize \centering LSTM (1 layer) & \footnotesize \centering 6 & \footnotesize \centering 25 & \footnotesize \centering 1 month & \footnotesize \centering 3 & \footnotesize \centering 64.8$\%$ & \footnotesize 64.3$\%$ \\
\hline
\end{tabular}
\label{tab:difdurantion}
\vspace{-0.25in}
\end{table}

\begin{table}
\centering
\caption{\small Impact of Different Learning Models}
\vspace{-0.1in}
\begin{tabular}{|p{1.1cm}|p{0.4cm}|p{0.7cm}|p{0.9cm}|p{0.55cm}|p{1.35cm}|p{0.9cm}|}
\hline\hline
\footnotesize \centering Model & \footnotesize \centering Time Slot & \footnotesize \centering Feature Size & \footnotesize \centering Window Size & \footnotesize \centering Step Nodes & \footnotesize \centering Accuracy (Validation) & \footnotesize Accuracy (Testing) \\ [0.1ex]
\hline
\footnotesize \centering LR & \footnotesize \centering 6 & \footnotesize \centering 25 & \footnotesize & \footnotesize & \footnotesize \centering 61.1$\%$ & \footnotesize 60.6$\%$ \\
\hline
\footnotesize \centering RF & \footnotesize \centering 6 & \footnotesize \centering 25 & \footnotesize & \footnotesize & \footnotesize \centering 64.3$\%$ & \footnotesize 63.7$\%$ \\
\hline
\footnotesize \centering CNN & \footnotesize \centering 6 & \footnotesize \centering 25 & \footnotesize & \footnotesize & \footnotesize \centering 65.9$\%$ & \footnotesize 65.6$\%$ \\
\hline
\footnotesize \centering LSTM (1 layer) & \footnotesize \centering 6 & \footnotesize \centering 25 & \footnotesize \centering 1 week & \footnotesize \centering 12 & \footnotesize \centering 69.4$\%$ & \footnotesize 68.7$\%$ \\
\hline
\footnotesize \centering LSTM (2 layers) & \footnotesize \centering 6 & \footnotesize \centering 25 & \footnotesize \centering 1 week & \footnotesize \centering 12 & \footnotesize \centering 77.6$\%$ & \footnotesize 77.2$\%$ \\
\hline
\footnotesize \centering LSTM (3 layers) & \footnotesize \centering 6 & \footnotesize \centering 25 & \footnotesize \centering 1 week & \footnotesize \centering 12 & \footnotesize \centering 73.1$\%$ & \footnotesize 72.5$\%$ \\
\hline
\end{tabular}
\label{tab:difmodel}
\vspace{-0.2in}
\end{table}

The results are shown in Table~\ref{tab:difdurantion} and \ref{tab:difmodel}. From the results, we can discover that there is a rapid accuracy increase from $k=3$ to $k=7$, but the improvement saturates as $k$ increases beyond 7 and in fact drops when $k=30$.
The results reveal that the step counts captured in a week contains sufficient information to allow us to forecast obesity status improvement at the end of the 3 months period. 
%The results indicate that the day-week-month variable contributes to the forecasting performance well, which means the history of every week in every time duration is sufficient to forecast future obesity status.
%It is because for the limited number of time step nodes, the models cannot learn enough relationships among different time series features. Whereas, the models will meet with vanishing gradient problems, where they will forget the learned information from the past when the number of nodes is large.  
This finding  coincides with a commonly used activity assessment method called the International Physical Activity Questionnaire (IPAQ) \cite{ipaq}, where the central question it has is ``The Time People Spent Being Physically Active in the Last 7 Days". 
%This consistency empirically ensures that the assessments' questions are reasonable for capturing participants' health status based on their activities. 

The CNN and LSTM-based deep learning models perform better  than the traditional learning models with the LSTM-based model performs 15\% better. However, the CNN model performs poorer than LSTM models since it lacks the capability to learn long-term correlations among time series features. 
In addition, the results also show that among all learning models, the 2-layer LSTM model has the highest accuracy. In this 2-layer LSTM model, the first layer acts as a feature extractor and the second layer learns the correlations among long-term dominant features.
The prediction performance drops when more than 2 layers are used for the LSTM-based model since overfitting begins to appear in this complicated model.

In addition, we have also separated the participants into 2 age-group (above/below 50 years old) and trained a 2-layer LSTM model for each age group. We obtain an accuracy of 80.7\% for those above 50 years old and an accuracy of 74.6\% for those below. This can be explained as follows: the weight loss of younger adults is also greatly influenced by their diets which is not captured in the application. Older adults typically have better diets and hence their weight loss can be more easily predicted using step counts.
% the effect of depth on the ability of LSTM to express correlation ranging over long time scales will decrease when the model becomes complex (e.g., with three layers).

\subsubsection{RQ3: How to interpret the learning model}

In order to better understand our learning scheme, we conduct Exp4 to explain the inner working of our 2-layer LSTM model and rationale behind its predictions.

\begin{figure}
\subfloat[\scriptsize Interpretation of the $10^{th}$ Time Step]{\includegraphics[height=1.5in,width=1.7in]{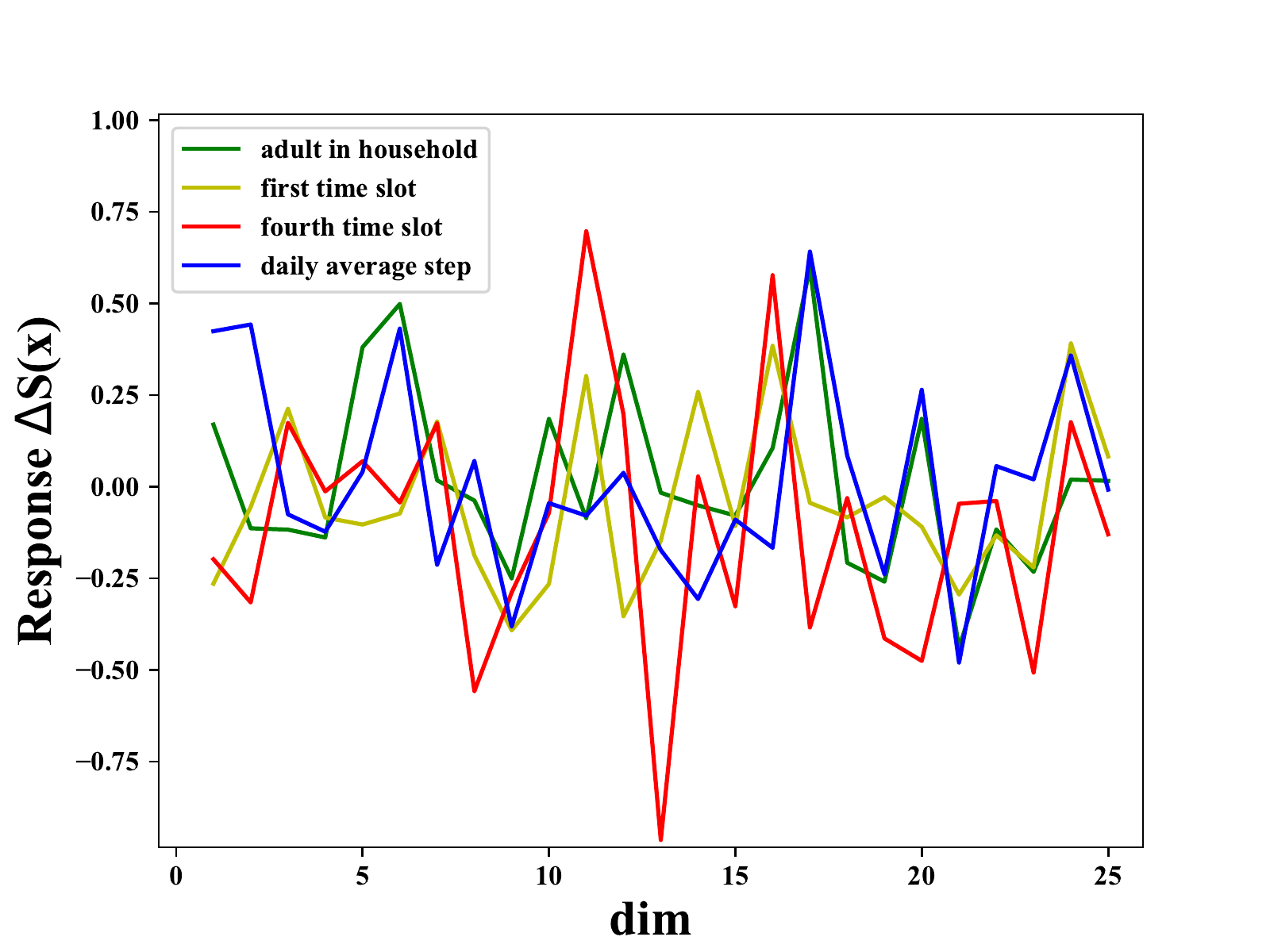}}
\hspace{0.05in}
\subfloat[\scriptsize Interpretation of the $12^{th}$ Time Step]{\includegraphics[height=1.5in,width=1.7in]{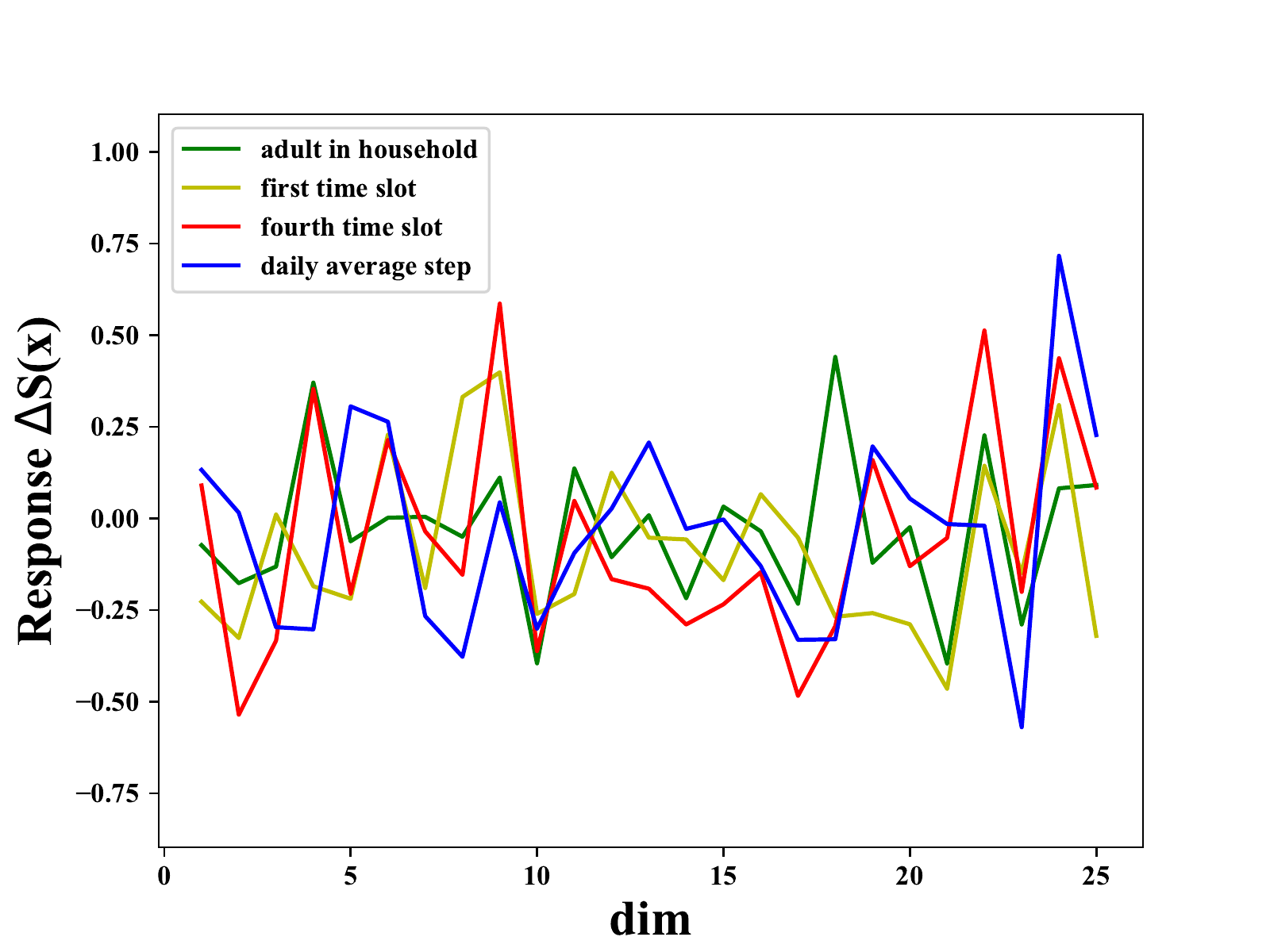}}
\vspace{-0.05in}
\caption{\small Two-layer LSTM Model Interpretation}
\label{fig:inter-1-2}
%\vspace{-0.25in}
\end{figure}

Exp4: Inspired by the idea of using interpretable representations to explain functions of network components, we propose a method to interpret the hidden state units using different inputs. 
Since our model typically has many-to-many relationships between hidden state units and inputs, we perform the following steps to study how different variables affect the various hidden state units: (i) we first choose four features (shown in Fig~\ref{fig:inter-1-2}), (ii) for each selected feature, we compute $S(X_t)$ values with all input features  as well as with modified features where that selected feature is zeroed out and record the differences in these two $S(X_t)$ values, (iii) based on the computed difference values, we determine which time steps has the largest impact on the prediction results. In this experiment, we set $m=6$ (each time slot equals 4 hours).
%where it can take a sequence of data and generate outputs at each time step, so in this experiment we first select essential time steps at the $2^{nd}$ LSTM layer by checking the accuracy of their prediction results.
%Then, we analyze the behavior of those chosen time steps and compute their responses according to the different inputs.
%For simplicity, we only modify a limited number of features for illustration, where one feature will be removed from the input at each time. 

\begin{table}
\centering
\caption{\small Step Counts at Different Time Slots}
\vspace{-0.1in}
\begin{tabular}{|p{2.4cm}|p{1.2cm}|p{2.4cm}|p{1.2cm}|}
\hline\hline
\footnotesize \centering Top5 Active Time Slots & \footnotesize \centering Total Steps & \footnotesize \centering Top5 Inactive Time Slots & \footnotesize Total Steps \\ [0.1ex]
\hline
\footnotesize \centering 12:00:00 - 13:00:00 & \footnotesize \centering 14527 & \footnotesize \centering 03:00:00 - 04:00:00 & \footnotesize 224 \\
\hline
\footnotesize \centering 13:00:00 - 14:00:00 & \footnotesize \centering 14221 & \footnotesize \centering 02:00:00 - 03:00:00 & \footnotesize 240 \\
\hline
\footnotesize \centering 14:00:00 - 15:00:00 & \footnotesize \centering 14061 & \footnotesize \centering 01:00:00 - 02:00:00 & \footnotesize 384 \\
\hline
\footnotesize \centering 11:00:00 - 12:00:00 & \footnotesize \centering 13860 & \footnotesize \centering 04:00:00 - 05:00:00 & \footnotesize 472 \\
\hline
\footnotesize \centering 15:00:00 - 16:00:00 & \footnotesize \centering 13528 & \footnotesize \centering 00:00:00 - 01:00:00 & \footnotesize 834 \\
\hline
\end{tabular}
\label{tab:diftimeslots}
\vspace{-0.25in}
\end{table}

After checking the prediction accuracy of each time step, we found two crucial time steps ( e.g., $10^{th}$ \& $12^{th}$) and further analyze their hidden units' responses. The hidden units' responses of these two time steps are shown in Fig~\ref{fig:inter-1-2} (a)(b).
From the results, we can see that the hidden state units in the left and right ends are more responsive/sensitive to the important feature ``daily average step" than the less important feature ``adult in household". 
We also find that different time slot features yield different impacts on the expected responses of hidden units, with the ``fourth time slot" having the most effect. This coincides with the information we see from the five highest hourly step counts reported by participants as shown in Table~\ref{tab:diftimeslots}. 
From the results, we can see that the most negative exercise time is the ``first time slot" (since few people will walk during 00:00:00 - 4:00:00) and the ``fourth time slot" (12:00:00 - 16:00:00) is the most active period. These results confirm that our model learns relevant information and allow us to infer the most relevant variables that can be used to help participants improve their BMI losses e.g., be more active during the ``fourth time slot".
%Thus, based on the results, we can probably recommend the most active time slot for the obese patients to do exercise, which in turn shows that it is useful in keeping users' motivation to log their histories voluntarily.

\subsubsection{Impact of Data Augmentation}

Since any deep learning model typically has excellent performance on a large dataset, we conduct Exp5 to check if our model will perform better when we use a larger dataset.

Exp5: Based on the results of the previous experiments, we first set $n=90$, $k=7$ and $m=6$, which provided the best performance.
Then, we apply an augmentation method to generate $D_2$, where instead of using the data from total 12 weeks, we only select $M$ ($M<12$) weeks' data to train our model.
In this experiment, (i) we select data from the $1^{st}$ and the $12^{th}$ weeks; (ii) we also select ($M-2$) weeks' data from  ($2^{nd}$- $11^{th}$) weeks' data; (iii) we feed the selected data into our 2-layer LSTM learning model.

\begin{table}
\centering
\caption{\small Impact of Different Data Sizes}
\vspace{-0.1in}
\begin{tabular}{|p{2.0cm}|p{1.3cm}|p{0.8cm}|p{1.4cm}|p{1.3cm}|}
\hline\hline
\footnotesize \centering Model & \footnotesize \centering \# of Time Steps & \footnotesize \centering Data Size & \footnotesize \centering Accuracy (Validation) & \footnotesize Accuracy (Testing) \\ [0.1ex]
\hline
\footnotesize \centering LSTM (2 layers) & \footnotesize \centering 11 & \footnotesize \centering 3230 & \footnotesize \centering 83.7\% & \footnotesize 83.4\% \\
\hline
\footnotesize \centering LSTM (2 layers) & \footnotesize \centering 10 & \footnotesize \centering 14535 & \footnotesize \centering 86.5\% & \footnotesize 86.1\% \\
\hline
\end{tabular}
\label{tab:difdatasize}
\vspace{-0.2in}
\end{table}

The results in Table~\ref{tab:difdatasize} show that the performance of our model improves by 8\% on average when we use a larger high-quality dataset. It is reasonable since modern nonlinear machine learning techniques get better performance with more higher quality data, especially for deep learning architecture. However, if we choose
 a small $M$ value (e.g., $M=3$), the accuracy drops since the gap between different time steps increases and there are fewer time steps which prevent our model from learning well the appropriate time correlations .

\section{Conclusions and Future Works}

\label{sec:conclude}
In recent years, new healthcare applications utilizing emerging smart
devices with embedded sensors for improving users' health have become popular.
In this paper, we have proposed a RNN based time-aware architecture to predict obesity status improvement using participants' data collected via wearables, e.g., blood pressures, step counts and their demographics. 
%We have also presented insights gained from analyzing the collected data regarding how activity patterns affect obesity status improvements. 
%We have also present the insights of activity pattern on obesity prediction and provide an in-depth understanding regarding internal processes of our trained model.
Our experimental results confirm that our framework can decently forecast obesity status improvement using users' activity and health measurement data. Furthermore, we have also provided some interpretations on how different variables affect our model.
Thus, we believe that our effort is a good step in understanding how collected activity data and physical health measurements can be utilized to predict users' health status improvement.
%aid clinical care, which provides empirical exploration of the predictions of RNN on chronic diseases using activity data.
As for the future work, we intend to apply our predictive model to infer improvements in hypertension or diabetes health statuses.
In addition, further prediction accuracy improvement can be made if we can integrate additional information such as participants' diets. 
%Meanwhile, we intend to apply our predictive model for other diseases such as hypertension and diabetes, which in turn can ease the burden on clinicians by providing timely and assistive recommendations.
%In addition, we also want to extend our analytic system to support the visualization for some specialized RNN-based models, such as memory networks or attention models. 

%\vspace{-0.1in}
\bibliographystyle{IEEEtran}
\bibliography{../bib/reference}
%\vspace{-3mm}

\end{document}